\documentclass[journal,comsoc]{IEEEtran}
\usepackage{xcolor}
\usepackage[T1]{fontenc}
\usepackage{cite}

\ifCLASSINFOpdf
   \usepackage[pdftex]{graphicx}
  \graphicspath{{../pdf/}{../jpeg/}}
  \DeclareGraphicsExtensions{.pdf,.jpeg,.png}
\else
  \usepackage[dvips]{graphicx}
  \graphicspath{{../eps/}}
  \DeclareGraphicsExtensions{.eps}
\fi

\usepackage{amsmath}
\usepackage[cmintegrals]{newtxmath}

\ifCLASSOPTIONcompsoc
  \usepackage[caption=false,font=normalsize,labelfont=sf,textfont=sf]{subfig}
\else
  \usepackage[caption=false,font=footnotesize]{subfig}
\fi
\usepackage{makecell}
\usepackage{amsmath}
\usepackage{bm}
\usepackage{multirow}
\usepackage{algorithm}
\usepackage{algpseudocode}
\usepackage{amsmath}
\usepackage{graphics}
\usepackage{epsfig}
\usepackage{booktabs}
\usepackage{diagbox}

\begin{document}

\title{Active Deep Densely Connected Convolutional Network for Hyperspectral Image Classification}

\author{Bing~Liu,~Anzhu~Yu,~Pengqiang~Zhang,~Lei~Ding, Wenyue~Guo~, ~Kuiliang~Gao, Xibing~Zuo
\thanks{This work was supported by the National Natural Science Foundation of China under Grant 4200011505,41201477.}\thanks{Bing~Liu*,~Anzhu~Yu*,~Pengqiang~Zhang,Wenyue~Guo~, Kuiliang~Gao, Xibing~Zuo are with the PLA Strategic Support Force Information Engineering University, Zhengzhou,450001, China. Bing~Liu and ~Anzhu~Yu contribute equally to this work. (e-mail: liubing220524@126.com).L. Ding is with the Department
of Information Engineering and Computer Science, University of Trento,
38123 Trento, Italy.}
}

\markboth{IEEE Journal of Selected Topics in Applied Earth Observations and Remote Sensing}%
{Shell \MakeLowercase{\textit{et al.}}: Bare Demo of IEEEtran.cls for IEEE Communications Society Journals}

\maketitle

\begin{abstract}
Deep learning based methods have seen a massive rise in popularity for hyperspectral image classification over the past few years. However, the success of deep learning is attributed greatly to numerous labeled samples. It is still very challenging to use only a few labeled samples to train deep learning models to reach a high classification accuracy. An active deep-learning framework trained by an end-to-end manner is, therefore, proposed by this paper in order to minimize the hyperspectral image classification costs. First, a deep densely connected convolutional network is considered for hyperspectral image classification. Different from the traditional active learning methods, an additional network is added to the designed deep densely connected convolutional network to predict the loss of input samples. Then, the additional network could be used to suggest unlabeled samples that the deep densely connected convolutional network is more likely to produce a wrong label. Note that the additional network uses the intermediate features of the deep densely connected convolutional network as input. Therefore, the proposed method is an end-to-end framework. Subsequently, a few of the selected samples are labelled manually and added to the training samples. The deep densely connected convolutional network is therefore trained using the new training set. Finally, the steps above are repeated to train the whole framework iteratively. Extensive experiments illustrates that the method proposed could reach a high accuracy in classification after selecting just a few samples.
\end{abstract}


\IEEEpeerreviewmaketitle
\section{Introduction}
\IEEEPARstart{H}{yperspectral} images (HSIs) are one of the most important data sources in the field of remote sensing \cite{cao2020hyperspectral}. It is of great significance for many earth observation applications to classify each pixel of HSIs into different classes. Therefore, HSI classification has been extensively studied. Detailed spectral information of HSIs could provide a basis for distinguishing different ground surface materials. Naturally, early studies focus on how to use spectral information to complete HSI classification \cite{liu2019deep}. For example, support vector machines (SVMs) \cite{5439692}, decision tree \cite{7064745}, sparse representation \cite{7433957}, gaussian process \cite{5204216}, and extreme learning machine \cite{6656874} have been heavily studied for HSI classification. The aforementioned supervised classifiers directly take spectral features as input and usually obtain a low classification accuracy.

Feature extraction is always considered as an effective method to improve the classification accuracy of HSIs \cite{BingLiu2017,9125900}. In this context, linear discriminant analysis, principal component analysis, independent component analysis, manifold learning are applied to spectral features \cite{8116758}. These feature extraction methods only considers spectral features and have no obvious effect on improving classification accuracy. In order to further improve the performance of HSI classification, neighborhood information of samples is introduced into the classification procedure. A common way to consider the influence of neighborhood information on classification results is texture feature extraction. For example, a 3-D Gabor feature-based collaborative representation approach is proposed for HSI classification in \cite{6866884}. A 3-D dense local binary pattern method is designed for HSI classification in \cite{7831381}. Local binary patterns, Gabor features and spectral features are input together into a extreme learning machine classifier \cite{7010879}. The combination of texture features and spectral features greatly improves the classification accuracy of HSIs. In addition, morphological filters are also used to extract structural features of HSIs. This kind of feature extraction method is known as morphological attribute profiles \cite{6945376}. The above feature extraction methods could improve the classification accuracy of HSIs. However, these methods need to design feature extraction rules manually. And in order to obtain good classification results, the parameters of different HSIs need to be adjusted carefully.

Deep learning could automatically mine features suitable for downstream tasks from data \cite{BingLiu2020}. In recent years, deep learning based methods have been widely used in HSI classification. There are two main problems in deep learning for HSI classification. One is the high dimension of HSI data, the other is the small number of labeled samples that can be obtained in HSIs. To deal with the first problem, the dimension of HSIs is first reduced, and then the reduced feature vector or image patch is input into a deep learning model to complete classification. The representative works following this idea are SAE \cite{6844831}, DBN \cite{7018910}, CNN \cite{yue2016a}, GAN \cite{8307247}. Although data dimensionality reduction could cope with the high-dimensional HSIs, a lot of detailed information is lost. In order to make better use of the rich spatial-spectral information in HSIs to improve the classification accuracy, 3D-CNN \cite{8685710,liu2018spectralspatial} and RNN \cite{8399509,liu2018spectral} are also used in HSI classification. These methods do not need dimension reduction preprocessing and have been widely studied. To deal with the second problem, some advanced deep learning structures are introduced into HSI classification, such as residual learning \cite{8061020}, dense network \cite{9050922,8784389}, cascade network structure \cite{8662780}, deep random forest \cite{8736489} and so on. These models greatly improve the classification accuracy of HSIs. However, lacking labeled training samples is still the key factor restricting the application of deep learning in HSI classification.

Defining an efficient training set is one of the most delicate phases for the success of HSI classification routines. Active learning is often designed to build effective sets of training by iteratively bettering the performance of the model through sampling. It has been extensively studied in HSI classification. In igeneral, active learning can be grouped into three main classes: committee learner-based approaches, margin sampling based approaches, class probability distribution based approaches \cite{8937036}. Committee learner-based approaches selects the samples showing maximal disagreement between the different classification models in the committee \cite{di2010multiview}. SVMs rely on a sparse representation of the training data, margin sampling based approaches aim at finding the pixels more likely to become support vectors \cite{mitra2004segmentation}. Class probability distribution based approaches use the estimation of posterior probabilities of class membership to rank the candidates \cite{rajan2008an}.

More recently, some works that combine active learning with deep learning have also been studied for HSI classification. Specifically, a unified deep network combined with active transfer learning are designed for HSI classification in \cite{deng2019active}. An active learning algorithm based on a weighted incremental dictionary learning is proposed for HSI classification in \cite{7568999}. An active learning process to initialize the salient samples on the HSI data are designed in \cite{8531707}. A method to combine a multiclass-level uncertainty active criterion with a stacked autoencoder is designed in \cite{li2016active}. 

Although the aforementioned active learning methods have achieved excellent performance. These methods require manual design of active learning strategies. In this paper, a novel active deep learning method is proposed for HSI classification. Different from the conventional active learning methods, the proposed method uses a neural network to predict the loss value of input samples. The predicted  loss value could be used to measure the importance of input samples. Therefore, we can select the samples that are more likely to be misclassified according to the loss value. Then, the selected samples are manually labeled and added into the training set. Note that the neural network used to predict the loss value takes the middle layer features of a deep densely connected convolutional network as the input. Therefore, the proposed method can be trained in an end-to-end manner, which greatly simplifies the procedure of active learning. In summary, the major contributions of this article can be abridged in the following ways:
\begin{itemize}
\item The deep densely connected convolutional network is designed for the classification of HSIs. The designed deep densely connected convolutional network derive from the classic DenseNet121, which enables us to reuse the classic network model. Note that the number of network layers used in this work is far more than the existing deep learning models for HSI classification. This not only saves the work of network design, but also proves that deep network model can be used to increase the accuracy of classification of an HSI assignment. 
\item An active deep learning framework is proposed in order to reduce the labeling cost of HSI classification and improve the classification accuracy. The proposed active deep learning framework adds an additional network to the designed classification network to predict the importance of the input samples. In this way, the proposed framework selects a few of samples to be labeled manually. Note that the selected samples are more likely to be confused. Therefore, adding the selected samples into the training set could greatly improve the classification performance.
\item Three HSI data sets with label information are used to evaluate the proposed active deep learning framework. The experimental results demonstrate that the proposed framework could achieve high classification accuracy with only a small number of labeled samples.
\end{itemize}

The remaining parts of this paper are as follows:  the proposed active deep learning framework is explained in detail in the Section 2, the presentation of corresponding analysis and experimental results are listed in the Section 3, this paper concludes with a couple of discussions in the Section 4.

\section{The proposed active deep learning framework}

In this section, we will first give the architecture of the deep densely connected convolutional network. Subsequently, we present the proposed active deep learning framework in detail.

\subsection{Deep densely connected convolutional network} 

Deep learning methods have led to a series of breakthroughs for image classification.  Recent researches reveal that network depth is of crucial importance for image classification. However, deep neural networks usually face the problem of model degradation, which makes the deep neural networks difficult to train. The extensive application of large-scale deep neural network model benefits from the proposal of skip-connection (e.g. residual learning \cite{he2016deep}). Motivated by residual learning, densely connected convolutional network introduces direct connections from any layer to all subsequent layers, which further improves the information flow between layers and the classification performance. Therefore, a deep densely connected convolutional network is used as the backbone network of this work.

\begin{figure*}[htbp]
\centering
\includegraphics[width=0.6\linewidth]{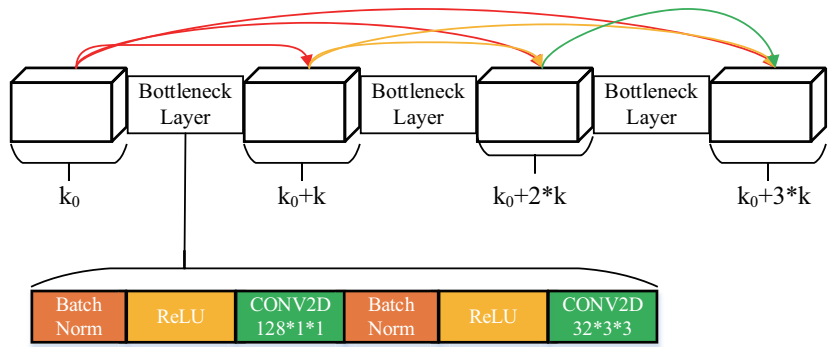}
\caption{Illustration of a standard dense block. CONV2D refers to a convolutional layer, BatchNorm refers to a batch normalization layer, ReLU refers to a ReLU layer.}
\label{fig1}
\end{figure*}

Dense block is the key component of a deep densely connected convolutional network. Fig. 1 shows the layout of a standard dense block that comprises three bottleneck layers. Each layer of a dense block is connected directly to the other layers in a way known as the skip-connection. For every layer, the preceding layers feature maps are considered to be separate inputs whereas its specific feature maps are passed on as inputs to all succeeding layers. Formally, the $l^{th}$ layer is defined as : 

\begin{equation}
\bm{x_l} = H_l([\bm{x_0}, \bm{x_1},\bm{x_2}, ......, \bm{x_{l-1}} ])
\end{equation}

\noindent Where the $\bm x$ values represent the concatenation of the feature-maps produced in layers $0, 1, ..., l-1$. The multiple inputs of $H^l(\cdot)$ in Eq. (1) are concatenated into a single tensor as the output of a dense block. In a dense block, the number of input feature maps is $k_0$. In each bottleneck layer, the number of output feature maps is fixed to $k$. Consequently, the number of output feature maps of this dense block is $k_0 + 3 \times k$ wher $k$ is the growth rate of a dense block. Following the oringinal paper, a bottleneck layer is a function of six consecutive operations: batch normalization (BatchNorm) \cite{ioffe2015batch}, a rectified linear unit (ReLU) \cite{nair2010rectified}, a $128 \times 1 \times 1$ convolution (CONV2D), BatchNorm, a ReLU and a $32 \times 3 \times 3$ CONV2D. In other word, the growth rate $k$ is 32.

As shown in Fig.2, deep densely connected convolutional network comprise one convolutional layer, one fully connected network layer, three transition layers, four dense blocks, and one pooling layer. ``CONV+BN'' denotes a convolutional layer followed by a BN layer, ``ReLU'' denotes a ReLU activation function, ``Max pool'' denotes a max pooling layer, ``Dense+Block'' denotes a dense block with several bottleneck layers, ``Transition'' denotes the transition layer, ``'GAP'' denotes the global average pooling layer. Transition layers use $1 \times 1$ convolution kernel to reduce the number of feature maps. In case a dense block outputs $m$ feature maps, we make the succeeding transition layer generate $m\theta$ feature maps, where $0 <\theta \leqslant 1$ is the compression factor.  

\begin{figure*}[!t]
\centering
\includegraphics[width=0.8\linewidth]{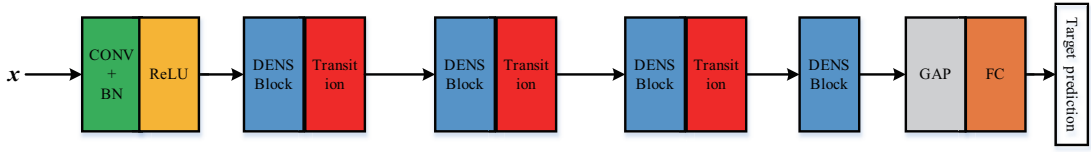}
\caption{Deep densely connected convolutional network structure. }
\label{fig2}
\end{figure*}

The original  DenseNet \cite{huang2017densely} paper provides DenseNet121, DenseNet169 and DenseNet201. The parameter configurations of DenseNet121, DenseNet169 and DenseNet201 are listed in Table 1.

\begin{table*}[!t]
\centering
\renewcommand{\arraystretch}{1.3}
\caption{Details of deep densely connected convolutional network used as the base classifier. Dense block (DB), global max pooling (GAP). }
\label{table1}
\begin{tabular}{@{}cccccccccccccc}
\toprule
Layers &DenseNet121&DenseNet169&DenseNet201 \\
\midrule
Convolution layer&\multicolumn{3}{c}{$64\times3\times 3$, stride 2} \\
\midrule
DB1&$\left[ \begin{array}{lcl} 128\times 1 \times 1 \\ 32\times 3\times 3 \end{array}\right] \times 6$&$\left[ \begin{array}{lcl} 128\times 1 \times 1 \\ 32\times 3\times 3 \end{array}\right] \times 6$&$\left[ \begin{array}{lcl} 128\times 1 \times 1 \\ 32\times 3\times 3 \end{array}\right] \times 6$\\
\midrule
Transition1&\multicolumn{3}{c}{\makecell{ $128 \times 1\times 1$  \\ $2\times2$ max pooling, stride 2 }}\\
\midrule
DB2&$\left[ \begin{array}{lcl} 128\times 1 \times 1 \\ 32\times 3\times 3\end{array}\right] \times 12$&$\left[ \begin{array}{lcl} 128\times 1 \times 1 \\ 32\times 3\times 3 \end{array}\right] \times 12$&$\left[ \begin{array}{lcl} 128\times 1 \times 1 \\ 32\times 3\times 3 \end{array}\right] \times 12$\\
\midrule
Transition2&\multicolumn{3}{c}{\makecell{ $256 \times 1\times 1$ \\ $2\times2$ max pooling, stride 2 }}\\
\midrule
DB3&$\left[ \begin{array}{lcl} 128\times 1 \times 1 \\ 32\times 3\times 3 \end{array}\right] \times 24$&$\left[ \begin{array}{lcl} 128\times 1 \times 1 \\ 32\times 3\times 3 \end{array}\right] \times 32$&$\left[ \begin{array}{lcl} 128\times 1 \times 1 \\ 32\times 3\times 3 \end{array}\right] \times 48$ \\
\midrule
Transition3&\multicolumn{3}{c}{\makecell{ $512 \times 1\times 1$\ \ \ \ \ \ \ \ \ \ \ \ \ \ \ \ \  $832 \times 1\times 1$\ \ \ \ \ \ \ \ \ \ \ \ \ \ \ \ \ \ \ $960 \times 1\times 1$ \\ $2\times2$ max pooling, stride 2 }}\\
\midrule
DB4 &$\left[ \begin{array}{lcl} 128\times 1 \times 1 \\ 32\times 3\times 3 \end{array}\right] \times 16$&$\left[ \begin{array}{lcl} 128\times 1 \times 1 \\ 32\times 3\times 3 \end{array}\right] \times 32$&$\left[ \begin{array}{lcl} 128\times 1 \times 1 \\ 32\times 3\times 3 \end{array}\right] \times 32$\\
\midrule
GAP&\multicolumn{3}{c}{$4 \times 4$}\\
\midrule
FC&\multicolumn{3}{c}{1000D fully-connected, softmax}\\
\bottomrule
\end{tabular}
\end{table*}

The original deep densely connected convolutional network is designed to identify the natural image in the Imagenet dataset. Its input size is $224 \times224 \times 3$. Different from natural images, HSIs consist of many bands. Following the idea of \cite{8116758}, different bands of HSI are input into the network as different feature maps. In this way, HSI cube can be input into the designed network without dimension reduction. In addition, a large number of studies show that considering the spatial neighborhood information in a neural network could increase the performance of HSI classification. Consequently, the input size of the deep densely connected convolutional network for HSI classification is $m \times m \times b$, where $b$ is the number of bands, $m$ is the size of neighborhood.

\subsection{Active deep densely connected convolutional network }

In the HSI classification task, we are able to collect a massive pool of samples $U_N$ that have not been labeled. The subscript $N$ represents the sum of samples to be classified. Then, $K0$ samples are randomly selected from the unlabeled pool. The selected samples are manually labeled as the initial labeled training dataset. Once an initially labeled training dataset is obtained, we can train a base classifier. Then we select a small number of unlabeled samples and request human to interpret them to add into the training dataset. The resulting training dataset is used for retraining the classifier. After several iterations, a small number of samples can be used to obtain a highpla classification accuracy. This is a standard active learning procedure. Note that active learning usually selects samples that are not easily distinguished to annotate.

The key of active learning is how to select representative samples to label. In order to make the designed deep densely connected convolutional network active learning, an additional network (loss prediction model) is added to the base classifier to predict the loss of samples. The larger the predicted loss value is, the more likely the sample will be misclassified by the classifier. These samples that are not easy to distinguish are the ones we need to select for manual annotation. The additional network for loss prediction is shown in Fig. 3. A global average pooling (GAP) layer is applied to the output features of the four dense block to obtain a one dimensional feature vector. A Fully-Connected ( FC ) layer is applied to the feature vector of different dense blocks to make the different features have the same dimension. The different features are added to input into a FC layer to predict the loss of the input sample.

As shown in Fig. 3, the final loss function comprises of dual parts. Formally, the final loss function is defined as:

\begin{equation}
Loss = Loss_{target}(\hat{\bm{y}}, \bm{y}) + Loss_{loss}(\hat{l},l)
\end{equation}

$Loss_{target}(\hat{\bm{y}}, \bm{y})$ is a standard cross-entropy loss function. With a sample $\bm{x}$, we could attain a class prediction $\hat{\bm{y}}$ through the deep densely connected convolutional network. $Loss_{target}$ is calculated from the known class $\bm{y}$ and the predicted class $\hat{\bm{y}}$. Formally, $Loss_{target}$ is defined as:

\begin{equation}
Loss_{target}(\hat{\bm{y}}, \bm{y}) = \sum^C_i y_i \log (\hat{y_i})
\end{equation}

\noindent where $y_i$ and $\hat{y_i}$ are the groundtruth and the deep densely connected convolutional network score for each $class_i$.

$Loss_{loss}(\hat{l},l)$ is the loss-prediction-loss-function. The MSE of the loss value perhaps is the most direct means to define the loss-prediction-loss-function. However, the value of the real loss $l$ declines in general with the continuous learning of the target model. In other words, the real value of $l$ is a variable. Optimizing the MSE of the loss value will make the additional network (loss prediction module) acclimate coarsely to the loss changes, instead of fitting to the precise value. This would lead to learning a bad loss prediction model. A bad loss prediction model can not select samples that are important for classification tasks. To discard the overall scale of $l$, the loss prediction loss-prediction-loss-function is computed by comparing a pair of samples. Mini-Batch Gradient Descent is used to optimize the proposed framework. Supposing the batch size is $B$, we could come up with $B/2$ sample pairs such as $\{ \bm{x}^p = (\bm{x}_i, \bm{x}_j)\}$. The subscript $p$ denotes that it is made up of two samples namely a sample pair. Note that $B$ should be an even number. The loss-prediction-loss-function is described as

\begin{equation}
\begin{split}
Loss_{loss}(\hat{l^p},l^p) = \max(0, -(\hat l_i - \hat l_j) \cdot \ell(l_i, l_j) + \xi)\\
s.t.\ \ \  \ell(l_i, l_j) = \begin{cases}
+1, if \ \ \  l_i > l_j\\
-1, \ \ \  otherwise\\
\end{cases}
\end{split}
\end{equation}

\noindent The subscript $p$ represents the pair of $(\bm{x}_i, \bm{x}_j)$, $\xi$ is a positive number, $l_i$ and $ l_j$ are the real loss of $\bm{x}_i$ and $\bm{x}_j$, $\hat l_i$ and $\hat l_j$ are the prediction loss of $\bm{x}_i$ and $\bm{x}_j$. When $l_i > l_j$, the loss-prediction-loss-function $Loss_{loss}$ states that no loss is given to the model only if $\hat l_i$ is larger than $\hat l_j + \xi$. If a loss is given to the model, the given loss would force it to increase $\hat l_i$ and decrease $\hat l_j$. In this way, the loss prediction model completely discard the overall scale changes \cite{yoo2019learning}.

\begin{figure*}[!t]
\centering
\includegraphics[width=0.9\linewidth]{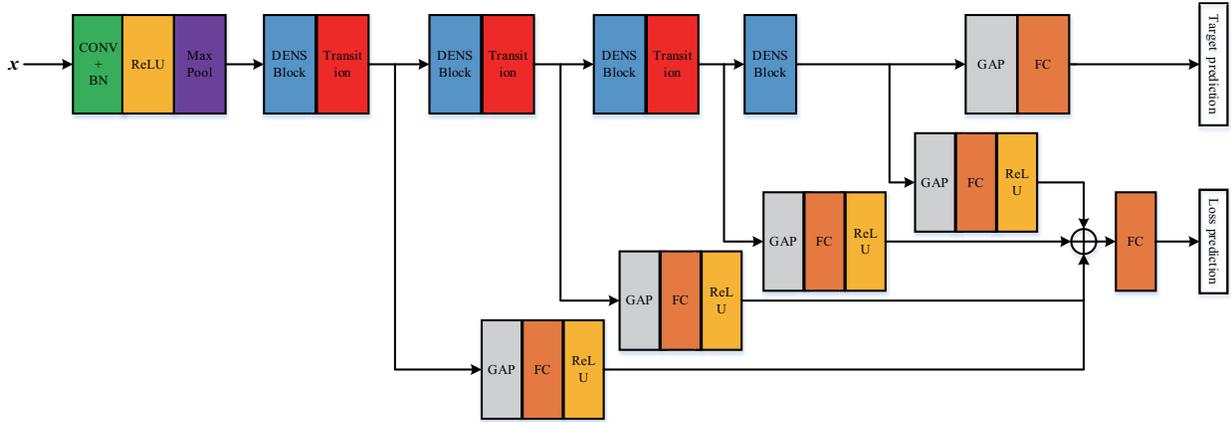}
\caption{Illustration of the active deep densely connected convolutional network. CONV+BN denotes that a convolutional layer is followed by a batch normalization layer, ReLU denotes a ReLU layer, DENS Block denotes a dense block, MaxPool and GAP denote the max pooling layer and the global average pooling layer respectively.}
\label{fig2}
\end{figure*}

To this end, the final loss function is computed as:

\begin{equation}
\frac 1 B \sum_{i=1}^B Loss_{target}(\bm{\hat y_i}, \bm{y_i}) + \frac 2 B \sum_{p=1}^{\frac B 2} Loss_{loss}(\hat{l^p},l^p)
\end{equation}

\noindent Minimizing this final loss function will make the prediction model learn to select the most informative samples and ask human oracles to annotate them for the next active learning stage.

\section{Experimental results and analysis}
The proposed active framework is implemented by the pytorch library. The experimental results are generated on a personal laptop equipped with an Intel Core i7-9750H with 2.6GHz and a Nvidia GeForce RTX 2070M.
\subsection{Experimental data sets}

In this section, three real HSI data sets including the University of Pavia, Indiana Pines and Salinas are used to demonstrate the effectiveness of the proposed active framework. The University of Pavia data set is acquired by the Reflective Optics Imaging Spectrometer System sensor. It has a geometric resolution of $1.3$ m. In this data set, 103 spectral bands could be used for classification.  The image size is $610\times 340$ pixels. 42776 pixels with nine classes are labeled.  The Indiana Pines data set is gathered by the Airbone Visible Infrared Imaging Spectrometer sensor and consists of $145\times145$ pixels. 200 bands could be used for classification. In this scene, 10249 pixels with sixteen classes are labeled.  The Salinas data set is also gathered by the Airbone Visible Infrared Imaging Spectrometer sensor and consists of $512\times217$ pixels. Its spatial resolution is 3.7 m. 204 bands could be used for classification. 54129 pixels with sixteen classes are labeled in this data set.

\subsection{Parameters setting and analysis}

The input size of the designed deep densely connected convolutional network is $32\times 32\times b$, where $b$ is the band number of HSIs. In general, training a CNN requires setting the learning rate, the number of epoch, the optimizer and the batch size. In this paper, the widely used Adam \cite{kingma2014adam} optimizer is used to optimize the proposed active deep learning framework. The batch size is set to be $10$, as the number of labeled samples for training is limited. The learning rate is set to be $0.001$ and the max training epoch is 200. This setting could not only ensure that the network is fully trained, but also ensure the stability of the training procedure.

First, 160 samples are randomly selected and manually labeled  to construct an initial training set. The proposed framework is trained by the initial training set. When the proposed framework is fully trained, the loss prediction model is used to compute the loss of the unlabeled samples. We rank the predicted loss of unlabeled samples, and select a small number of samples to label manually according to the loss value. The selected samples are then added into the training set. The proposed framework is trained by the new training set. In this work, 10 samples are selected in each iteration in order to reduce the cost of manual labeling. 32 iterations training are conducted in order to observe the effectiveness of the proposed method. In other words, the total number of labeled samples used for training is 480.

The classification accuracy with different backbone networks are shown in Fig. 4. The backbone networks include ResNet18, ResNet34, Resnet50, ResNet101, Resnet152, DenseNet121, DenseNet169 and DenseNet201. From the results of Fig. 4, these deep networks could obtain ideal classification results through the active learning strategy proposed in this paper. However, the classification accuracy of the DenseNets is generally higher than that of the ResNets. This proves the rationality of using dense network in this paper. Considering that using more layers will increase the training time, this paper uses DenseNet121 as the base classifier. 

\begin{figure*}
\begin{center}
\resizebox*{0.9\linewidth}{!}{\includegraphics{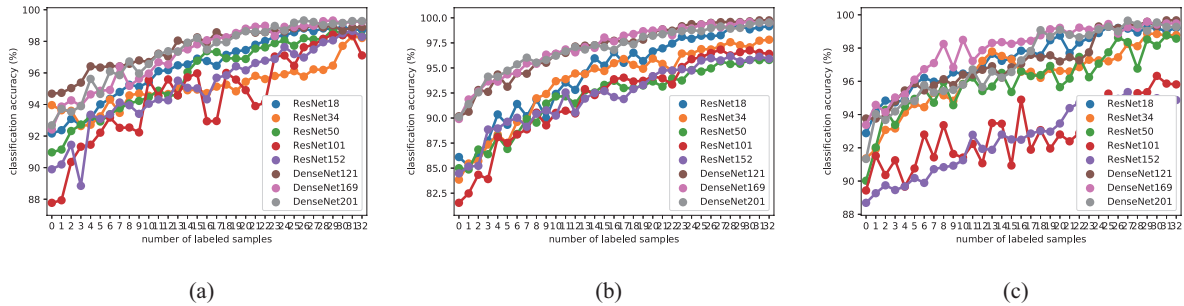}}\hspace{5pt}
\caption{The classification accuracy (OA) with different backbone networks. (a) University of Pavia data set (b) Indiana Pines data set (c) Salinas data set. }
\label{Figure4}
\end{center}
\end{figure*}

\subsection{Comparison results with the active learning methods}

In this section, the proposed active learning strategy is compared with the max entropy \cite{modAL2018} active learning method. To demonstrate the effectiveness of the proposed method, the max entropy active learning methods are conducted on three classifiers. The classification accuracy (OA) with different methods are shown in Fig. 5. In Fig. 5,  `Active+DenseNet121` denotes the proposed active learning method, `MaxEntropy+DenseNet121` denotes the combination of max entropy active learning and DenseNet121, `MaxEntropy+LeNet` denotes the combination of max entropy active learning and LeNet, `MaxEntropy+EMP` denotes the combination of max entropy active learning, EMP features and SVM. First, it could be found that the OA of `MaxEntropy+DenseNet121` is much higher than that of `MaxEntropy+LeNet` and `MaxEntropy+EMP`. This shows that reusing the classical DenseNet121 structure can greatly improve the classification performance. More importantly, the proposed active learning could outperform the classic max entropy method. In this work, the improvement of HSI classification accuracy is mainly due to two aspects, one is the improvement of network structure, the other is the proposed active learning strategy.

\begin{figure*}
\begin{center}
\resizebox*{0.9\linewidth}{!}{\includegraphics{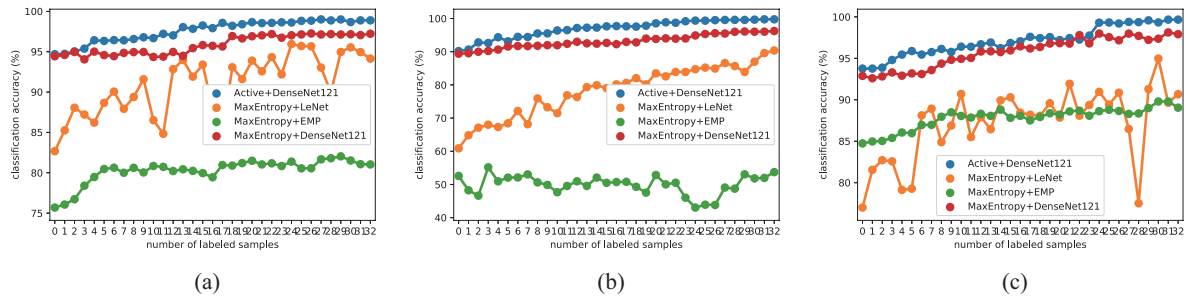}}\hspace{5pt}
\caption{The classification accuracy (OA) with different methods. (a) University of Pavia data set (b) Indiana Pines data set (c) Salinas data set. }
\label{Figure5}
\end{center}
\end{figure*}

\subsection{Comparison results with the state-of-the-art methods}

In this section, the performance of the proposed method (Active+DenseNet121) is compared with several state-of-the-art methods. The compared methods are listed as below:
\begin{enumerate}
\item EMP+SVM \cite{mura2010} is a spatial-spectral feature method for HSI classification. EMP features are extracted by repeating the opening operation and closing operation on a band image. Square structure element is used in the opening operation and closing operation. The radius of structuring elements are set to be 1,3,5,7,9, respectively. The optimal hyperplane parameters of the SVM classifier are determined  by five-fold cross validation. 
\item JCR (joint collaborative representation )\cite{6779644} investigates the relationship of hyperspectral neighbors based on nearest regularized subspace (NRS) classifier for HSI classification. It obtains a ideal classification result with only a few labeled samples.
\item 3D-CNN \cite{8685710} is a classical method for HSI classification. It could outperform the traditional machine learning methods such as SVM, random forest.
\item 3DCAE \cite{8694830} is an unsupervised spatial–spectral feature learning method based on 3D convolutional autoencoder. It is very effective in extracting spatial–spectral features. The parameters are set the same as the paper.
\item CNN-PPF \cite{7736139} use a novel pixel-pair method to significantly increase the number of labeled samples, ensuring that the advantage of deep CNN can be actually offered. This strategy greatly improve the HSI classification performance.
\item S-CNN+SVM\cite{8116758} is a supervised feature extraction method. It trains a siamese convolutional neural network (S-CNN) to increase the separability of different classes. Afterward, features extracted via S-CNN are used to train a linear SVM classifier. This method could increase the number of labeled samples, thus improving the classification.
\item DFSL+SVM \cite{liu2019deep} introduces meta learning method into HSI classification. The deep 3D-CNN is trained by the episode based meta learning method. Features extracted by the fully trained 3D-CNN could improve the HSI classification performance.
\end{enumerate}
 
200 labeled samples per class are randomly selected to train the compared methods. As for the compared methods, $N=200 \times C$ labeled samples are used as training set, where $C$ is the number of classes, $N$ is the total number of labeled samples used as training set. For example, 1800 labeled samples are used as the training set in the University of Pavia data set.  Note that there are 16 different landcover classes in the original ground truth of the Indiana Pines data set. However, only nine classes are used so as to avoid a few classes that have very few training samples \cite{7736139}. To demonstrate that the proposed method could reduce the labeling cost of HSI classification, only 480 labeled samples are used to train the proposed method (Active DenseNet121). Note that these samples are selected according to the predicted loss and manually labeled.

The class-specific accuracy, overall accuracy (OA), average accuracy (AA) and $\kappa$ of different methods for three HSI data sets are listed in Tables \uppercase\expandafter{\romannumeral2}-\uppercase\expandafter{\romannumeral4}. From these results, it could be found that EMP+SVM, JCR, 3D-CNN, 3DCAE, CNN-PPF, GCN, DFSL+SVM both obtain high classification accuracy. It is worth noting that the proposed method Active+DenseNet121 achieves a higher overall classification accuracy and uses less labeled samples. For example, in the Salinas data set, 3200 labeled samples are used to train the compared methods. In contrast with the compared methods, only 480 labeled samples are used to train the proposed active deep learning method. This shows that the proposed method can reduce the labeling cost of HSI classification under the premise of ensuring the classification performance.

In order to better observe the classification results, the classification maps of different methods on three HSI data set are shown in Figs. 6-8. To facilitate comparison between different methods, the ground truth maps are shown Figs. 6-8. From these maps, it could be found that the maps produced by the Active+DenseNet121 are highly consistent with the ground truth maps. For example, in the Salinas data set, there are more classification noises in Grapes\_untrained and Vinyard\_untrained categories of the maps generated by the comparison algorithm. This further proves the effectiveness of the proposed method.

\begin{table*}
\caption{Class-specific accuracy, OA, AA and $\kappa$ of different methods for the University of Pavia data set (Bold values represent the best accuracy among these methods in each case).}
\centering
{\begin{tabular}{|c|c|c|c|c|c|c|c|c|c|}
\hline
\makecell{Class\\No.}&\makecell{EMP+\\SVM}&JCR&\makecell{3D-\\CNN}&3DCAE&\makecell{CNN\\-PPF}&\makecell{S-CNN\\+SVM}&\makecell{DFSL\\+SVM}&\makecell{Active+\\DenseNet121}\\ 
\hline
1&93.27&97.15&99.03&92.87&97.23&97.63&97.18&99.77\\
2&95.79&98.60&98.11&97.46&95.27&99.38&99.40&100.0\\
3&91.14&97.52&88.56&91.90&95.13&96.71&97.90&99.95\\
4&99.22&99.41&83.51&97.68&96.89&99.22&98.40&95.95\\
5&99.41&100.0&99.49&99.85&99.99&100.0&100.0&100.0\\
6&95.63&98.33&95.33&98.65&98.55&97.69&99.56&99.40\\
7&97.74&98.65&96.31&97.74&96.56&97.52&99.25&96.92\\
8&89.63&92.67&97.58&86.01&94.43&95.55&95.52&99.76\\
9&100.0&99.58&96.25&99.26&99.39&100.0&99.68&90.92\\
\hline
OA (\%)&95.14&97.90&96.37&95.77&97.63&98.42&98.62&\textbf{99.28}\\
AA (\%)&95.76&97.99&94.82&95.71&97.04&98.19&\textbf{98.54}&98.07\\
$\kappa$ &93.60&97.23&95.02&94.42&96.90&97.90&98.17&\textbf{99.05}\\
\hline
N&1800&1800&1800&1800&1800&1800&1800&480\\
\hline
\end{tabular}}
\label{table2}
\end{table*}

\begin{table*}
\caption{Class-specific accuracy, OA, AA and $\kappa$ of different methods for the Indiana Pines data set (Bold values represent the best accuracy among these methods in each case).}
\centering
{\begin{tabular}{|c|c|c|c|c|c|c|c|c|c|}
\hline
\makecell{Class\\No.}&\makecell{EMP+\\SVM}&JCR&\makecell{3D-\\CNN}&3DCAE&\makecell{CNN\\-PPF}&\makecell{S-CNN\\+SVM}&\makecell{DFSL\\+SVM}&\makecell{Active+\\DenseNet121}\\
\hline
1&88.94&96.50&81.93&88.31&92.99&94.61&98.32&99.79\\
2&96.87&99.52&93.25&92.65&96.66&97.59&99.76&99.28\\
3&98.76&100.0&96.69&99.17&98.58&97.72&100.0&100.0\\
4&99.86&99.59&97.26&98.49&100.0&100.0&100.0&100.0\\
5&100.0&100.0&100.0&100.0&100.0&100.0&100.0&100.0\\
6&93.12&96.81&91.05&90.23&96.24&95.58&97.84&99.69\\
7&88.64&96.37&85.74&79.23&87.80&95.03&95.93&99.80\\
8&96.29&100.0&96.29&94.44&98.98&98.65&99.66&99.16\\
9&99.53&99.68&99.92&96.84&99.81&99.92&99.76&99.92\\
\hline
OA (\%)&93.95&98.04&91.23&90.03&94.34&96.95&98.35&\textbf{99.75}\\
AA (\%)&95.78&98.72&93.57&93.26&96.78&97.68&99.03&\textbf{99.74}\\
$\kappa$ &92.92&97.70&89.76&88.38&93.97&96.42&98.07&\textbf{99.71}\\
\hline
N&1800&1800&1800&1800&1800&1800&1800&480\\
\hline
\end{tabular}}
\label{table2}
\end{table*}

\begin{table*}
\caption{Class-specific accuracy, OA, AA and $\kappa$ of different methods for the Salinas data set (Bold values represent the best accuracy among these methods in each case).}
\centering
{\begin{tabular}{|c|c|c|c|c|c|c|c|c|c|}
\hline
\makecell{Class\\No.}&\makecell{EMP+\\SVM}&JCR&\makecell{3D-\\CNN}&3DCAE&\makecell{CNN\\-PPF}&\makecell{S-CNN\\+SVM}&\makecell{DFSL\\+SVM}&\makecell{Active+\\DenseNet121}\\
\hline
1&99.40&100.0&99.94&99.85&99.84&99.90&100.0&100.0\\
2&97.70&99.87&85.45&100.0&99.77&99.70&99.97&99.87\\
3&99.70&99.95&100.0&99.14&98.11&99.85&100.0&100.0\\
4&99.64&99.78&99.77&99.93&99.57&100.0&99.86&99.86\\
5&98.25&99.29&99.96&99.59&98.54&99.70&100.0&99.96\\
6&99.90&100.0&100.0&100.0&99.92&99.77&100.0&99.72\\
7&99.39&99.66&99.60&99.92&99.96&99.61&100.0&100.0\\
8&85.31&96.46&99.31&86.52&89.11&87.18&91.67&99.93\\
9&99.61&99.47&99.97&99.76&99.69&99.60&99.69&99.02\\
10&97.35&99.79&99.41&99.76&97.78&99.36&99.79&100.0\\
11&99.63&100.0&100.0&99.81&99.33&98.60&100.0&100.0\\
12&100.0&100.0&100.0&100.0&100.0&100.0&100.0&98.39\\
13&99.67&100.0&100.0&100.0&99.67&100.0&100.0&93.67\\
14&98.13&98.50&100.0&100.0&98.75&98.79&100.0&99.72\\
15&88.50&98.24&90.36&94.48&89.99&92.17&97.01&100.0\\
16&99.28&99.89&85.93&99.94&99.07&100.0&99.94&100.0\\
\hline
OA (\%)&94.92&98.85&97.28&96.34&94.87&96.06&97.81&\textbf{99.67}\\
AA (\%)&97.72&99.43&97.48&98.67&98.07&98.39&99.25&\textbf{99.38}\\
$\kappa$ &94.35&98.71&96.95&95.93&94.04&95.61&97.56&\textbf{99.63}\\
\hline
N&3200&3200&3200&3200&3200&3200&3200&480\\
\hline
\end{tabular}}
\label{table2}
\end{table*}

\begin{figure*}
\begin{center}
\resizebox*{0.8\linewidth}{!}{\includegraphics{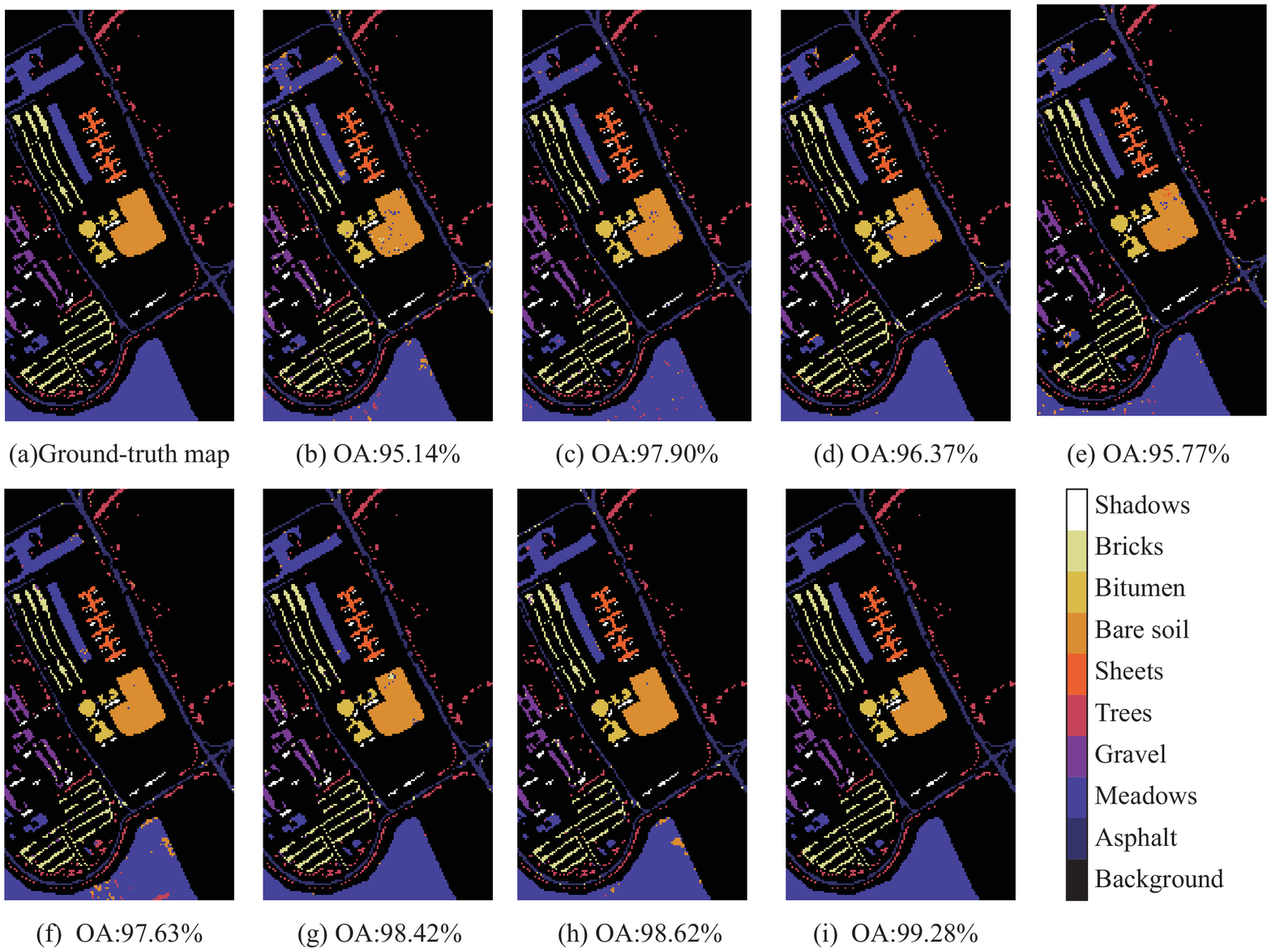}}\hspace{5pt}
\caption{Classification maps resulting from different methods for the University of Pavia data set. (a) Ground-truth map (b) EMP+SVM (c) JCR (d) 3D-CNN (e) 3DCAE (f) CNN-PPF (g) Resnet50 (h) DFSL+SVM (i) Active+DenseNet121. }
\label{Figure7}
\end{center}
\end{figure*}

\begin{figure*}
\begin{center}
\resizebox*{0.95\linewidth}{!}{\includegraphics{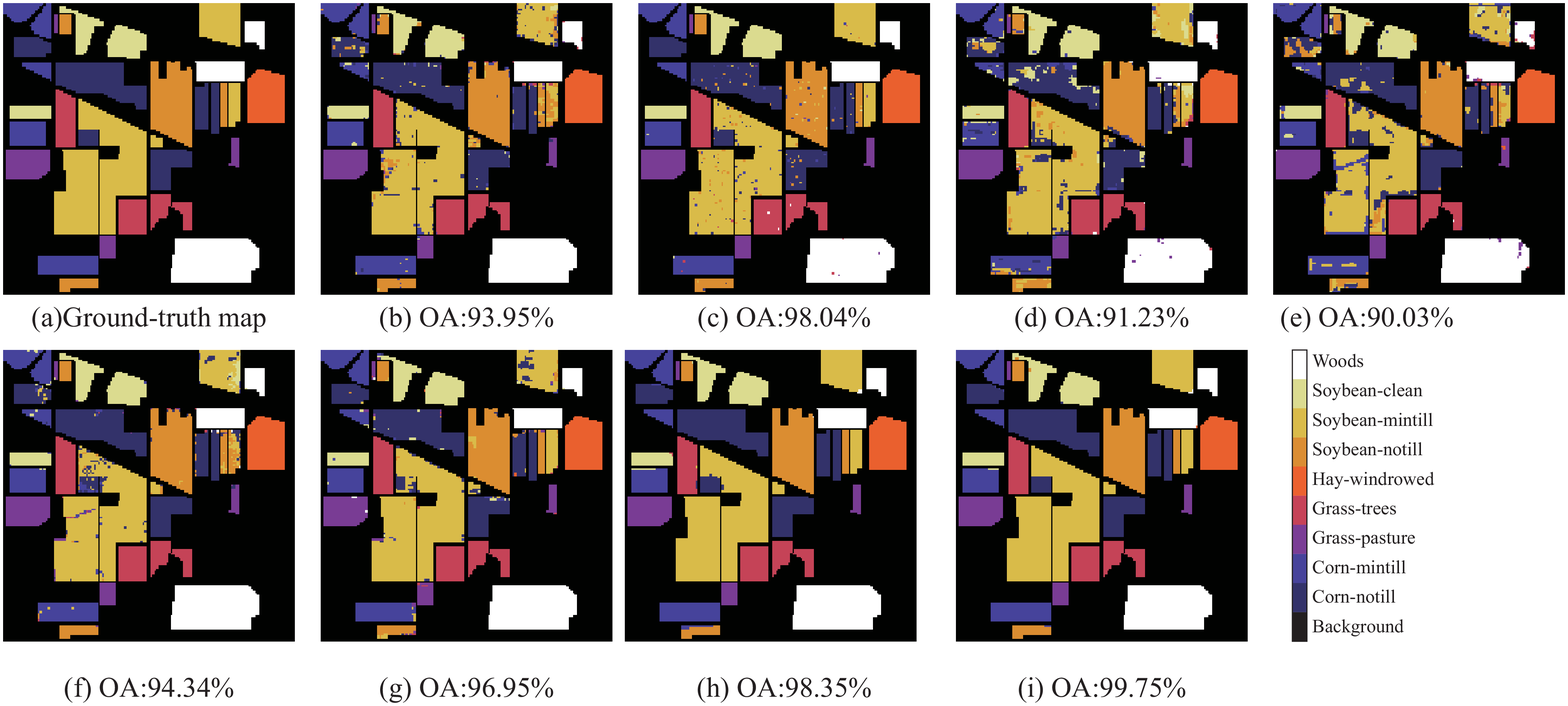}}\hspace{5pt}
\caption{Classification maps resulting from different methods for the Indiana Pines data set. (a) Ground-truth map (b) EMP+SVM (c) JCR (d) 3D-CNN (e) 3DCAE (f) CNN-PPF (g) Resnet50 (h) DFSL+SVM (i) Active+DenseNet121. }
\label{Figure8}
\end{center}
\end{figure*}

\begin{figure*}
\begin{center}
\resizebox*{0.8\linewidth}{!}{\includegraphics{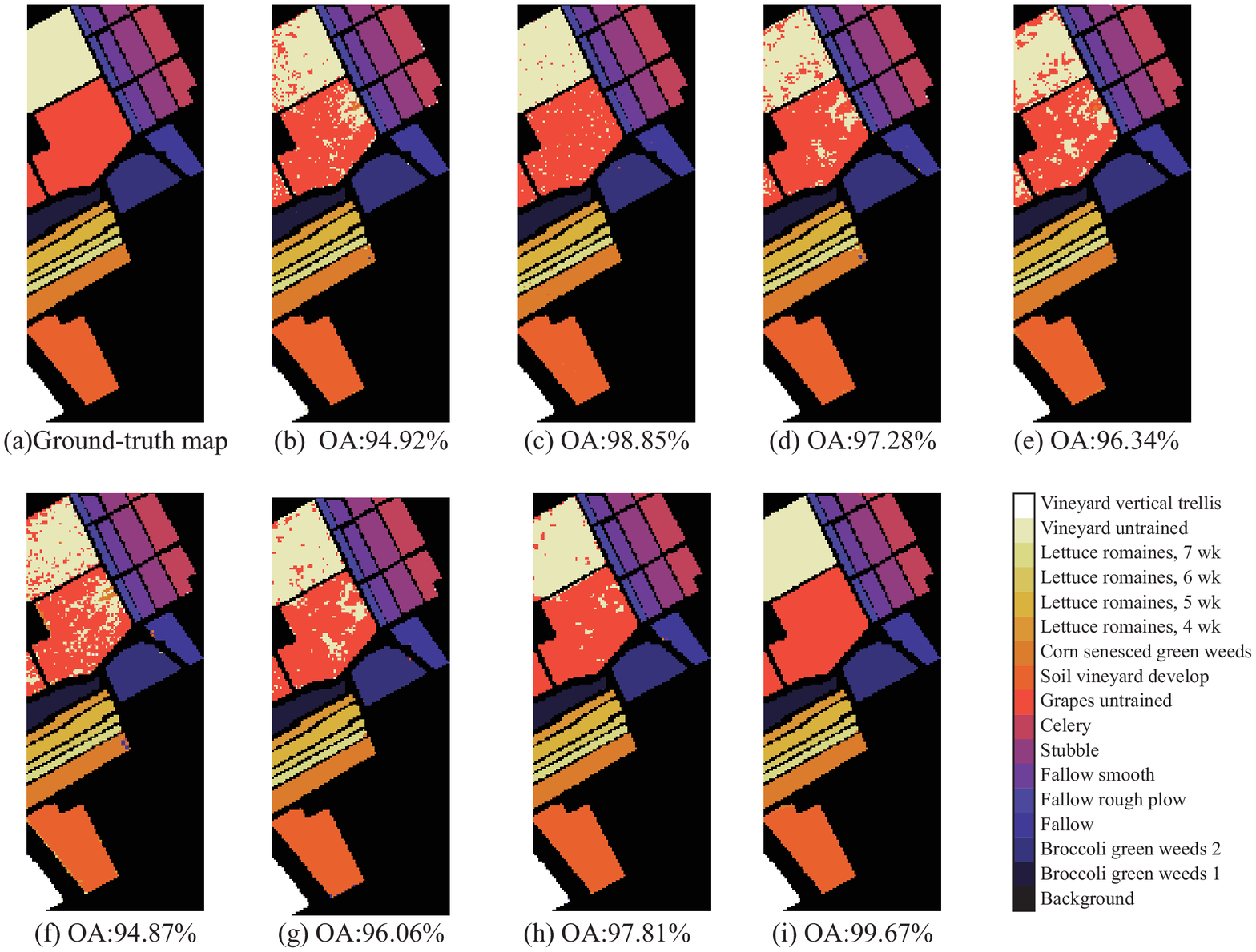}}\hspace{5pt}
\caption{Classification maps resulting from different methods for the Salinas data set. (a) Ground-truth map (b) EMP+SVM (c) JCR (d) 3D-CNN (e) 3DCAE (f) CNN-PPF (g) Resnet50 (h) DFSL+SVM (i) Active+DenseNet121. }
\label{Figure9}
\end{center}
\end{figure*}

\section{Conclusion}
In this paper, an active deep learning framework is proposed for HSI classification. The proposed framework consists of a base deep densely connected convolutional network classifier and a prediction model. The base deep densely connected convolutional network classifier is used to classify the input samples. The prediction model is used to predict the loss value of the input samples. Samples with large loss values are then selected for manual marking. Extensive experiments show that the proposed method can use less labeled samples to achieve higher classification accuracy, thus reducing the cost of labeling samples.

\section*{Acknowledgment}

We thank Prof. Mei Shaohui for providing the codes of 3DCAE.

\ifCLASSOPTIONcaptionsoff
  \newpage
\fi

\bibliography{report}   
\bibliographystyle{ieeetr}   

\end{document}